\documentclass{article}

\usepackage{microtype}
\usepackage{graphicx}
\usepackage{subcaption}
\usepackage{booktabs}
\usepackage{hyperref}
\usepackage{amsmath}
\usepackage{amssymb}
\usepackage{mathtools}
\usepackage{amsthm}
\usepackage[capitalize,noabbrev]{cleveref}

\usepackage[accepted]{icml2026}

\theoremstyle{plain}
\newtheorem{theorem}{Theorem}[section]

\theoremstyle{definition}
\newtheorem{definition}[theorem]{Definition}

\theoremstyle{remark}

\newcommand{\ours}{\textsc{TDGA}}
\icmltitlerunning{Directional Alignment Mitigates Reward Hacking in Reinforcement Learning for Language Models}

\begin{document}

\twocolumn[
  \icmltitle{Directional Alignment Mitigates Reward Hacking in Reinforcement Learning for Language Models}
\icmlsetsymbol{int}{$\diamond$}
\begin{icmlauthorlist}
  \icmlauthor{Wenlong Deng}{yyy,sch,amz,int}
  \icmlauthor{Jiaji Huang}{amz}
  \icmlauthor{Kaan Ozkara}{amz}
  \icmlauthor{Yushu Li}{yyy,sch}
  \icmlauthor{Christos Thrampoulidis}{yyy,sch}
  \icmlauthor{Xiaoxiao Li}{yyy,sch}
  \icmlauthor{Youngsuk Park}{amz}
\end{icmlauthorlist}

 \icmlcorrespondingauthor{Wenlong Deng}{dwenlong@ece.ubc.ca}

\icmlaffiliation{yyy}{University of British Columbia}
\icmlaffiliation{sch}{Vector Institute}
\icmlaffiliation{amz}{Amazon}

  \icmlkeywords{reward hacking, reinforcement learning, language models, optimization dynamics}

  \vskip 0.3in
]

\printAffiliationsAndNotice{\textsuperscript{$\diamond$}Work done during internship at Amazon. \textbf{This manuscript presents a preliminary study and is shared to encourage early discussion and feedback from the community.}}

\begin{abstract}
Reward hacking arises when a model improves a proxy reward by exploiting shortcuts rather than solving the intended task. We study this failure mode through the geometry of reinforcement learning updates in language models and argue that hacking emerges when optimization drifts away from a stable low-dimensional learning trajectory. We analyze this drift through dominant singular directions of parameter updates and show that reward-hacking runs exhibit substantially larger directional change than clean runs. Motivated by this observation, we introduce trusted-direction projection, which constrains gradients to remain within a clean reference subspace. Across reward-hacking experiments on mathematical reasoning, the proposed approach delays shortcut exploitation and better preserves task performance.
\end{abstract}

\section{Introduction}
Reinforcement learning (RL) has become a widely used approach for improving the reasoning capabilities of large language models (LLMs)~\citep{guo2025deepseek,deng2025effect}. However, RL training can suffer from reward hacking~\cite{skalse2022defining}, where a model achieves high reward by exploiting unintended shortcuts in the training environment rather than genuinely solving the target task~\cite{wang2025thinking,li2025generalist}. This failure mode is particularly concerning for LLM reasoning, as the proxy reward may indicate improvement even while true task performance degrades. For instance, when a dataset or evaluation pipeline contains exploitable artifacts~\cite{wang2025thinking}, the model may learn to depend on these artifacts instead of developing the intended reasoning ability.
\\
Prior work has largely framed reward hacking as a problem of reward misspecification~\cite{turpin2025teaching}. From this perspective, failures arise because reward functions or learned reward models do not fully capture the true objective, allowing models to optimize proxy signals in unintended ways. Existing approaches therefore focus on improving reward modeling~\cite{turpin2025teaching,li2025generalist,he2025gardo} or introducing additional regularization toward a reference model~\cite{laidlaw2024correlated}. While these methods are valuable, they face a fundamental limitation: constructing a perfect reward model is inherently difficult, particularly for complex reasoning tasks where the true objective is only partially specified. Strong regularization can also constrain the model's ability to learn beyond the reference policy.
\\
Recent work~\cite{ackermann2026gradient} has begun to investigate the learning mechanisms underlying reward hacking, showing that it is closely associated with sharp local minima and can therefore be mitigated by smoothing the optimization landscape~\cite{kwon2021asam}. However, these approaches primarily regulate the magnitude of parameter updates, without explicitly enforcing their directional alignment with the true objective. Meanwhile, RL updates in language models appear to exhibit a striking linear structure: a large fraction of the performance gain is captured by the leading singular direction of the parameter update matrix, and this dominant direction evolves along an approximately linear trajectory throughout training~\cite{cai2025predictability}. Building on this observation, we study reward hacking through the lens of optimization dynamics. We argue that reward hacking is not merely a consequence of imperfect reward design, but more fundamentally arises when gradient updates drift away from the model's intrinsic learning trajectory and enter directions that improve proxy reward while remaining misaligned with true task performance. 
\\
To mitigate reward hacking, we propose trusted-direction gradient alignment (\ours), which constructs a reliable optimization subspace by applying SVD to the parameter changes induced by a small number of clean supervised training steps. During RL training, we project gradients onto this trusted learning subspace, constraining updates to remain within a safer region of the parameter space. Our contributions are threefold:
\\
\noindent$\bullet$ We characterize reward hacking as directional drift in the dominant singular subspace of RL updates. 
\\
\noindent$\bullet$ We empirically show that clean training preserves directional consistency, whereas reward-hacking runs exhibit sharp rotations away from trusted directions. 
\\
\noindent$\bullet$ We introduce a trusted-direction gradient alignment that anchors RL updates to a clean reference subspace and substantially delays reward hacking.

\section{Related Work}\label{sec:related_work}
\textbf{Reward hacking in language-model RL.}
Reward hacking occurs when an agent achieves high proxy reward by exploiting a mismatch between the reward signal and the intended objective~\citep{skalse2022defining}. Existing mitigations mainly improve the reward signal itself, through stronger reward models~\citep{li2025generalist,liu2025inference}, task-specific anti-hacking schemes~\citep{he2025gardo}, or formal treatments of correlated proxies~\citep{laidlaw2024correlated}. However, perfect reward specification is often difficult. A complementary line of work studies hacking through optimization geometry: gradient regularization smooths unstable updates~\citep{ackermann2026gradient}, while sharpness-aware methods favor flatter and more robust minima~\citep{foret2020sharpness,kwon2021asam}. These methods control local smoothness or update magnitude, but do not explicitly preserve alignment with task-relevant learning directions.
\\
\textbf{Optimization geometry and learning dynamics.}
The learning dynamics of LLMs have recently received increasing attention. For example, \cite{deng2025efficient} studies learning dynamics for identifying valuable fine-tuning data, while \cite{deng2025effect,deng2025grpo} analyze likelihood dynamics to diagnose RL training collapse. Recent work on language-model RL further shows that parameter updates often exhibit a low-rank and approximately linear structure, where the leading singular directions explain much of the performance change induced by training~\citep{cai2025predictability}. Our work connects these lines of research by interpreting reward hacking as directional drift away from a trusted low-dimensional learning trajectory, and by projecting RL gradients back onto that trajectory.

\section{Dominant Update Directions}\label{sec:d_direct}
\begin{figure*}[t]
    \centering
    \includegraphics[width=0.9\linewidth]{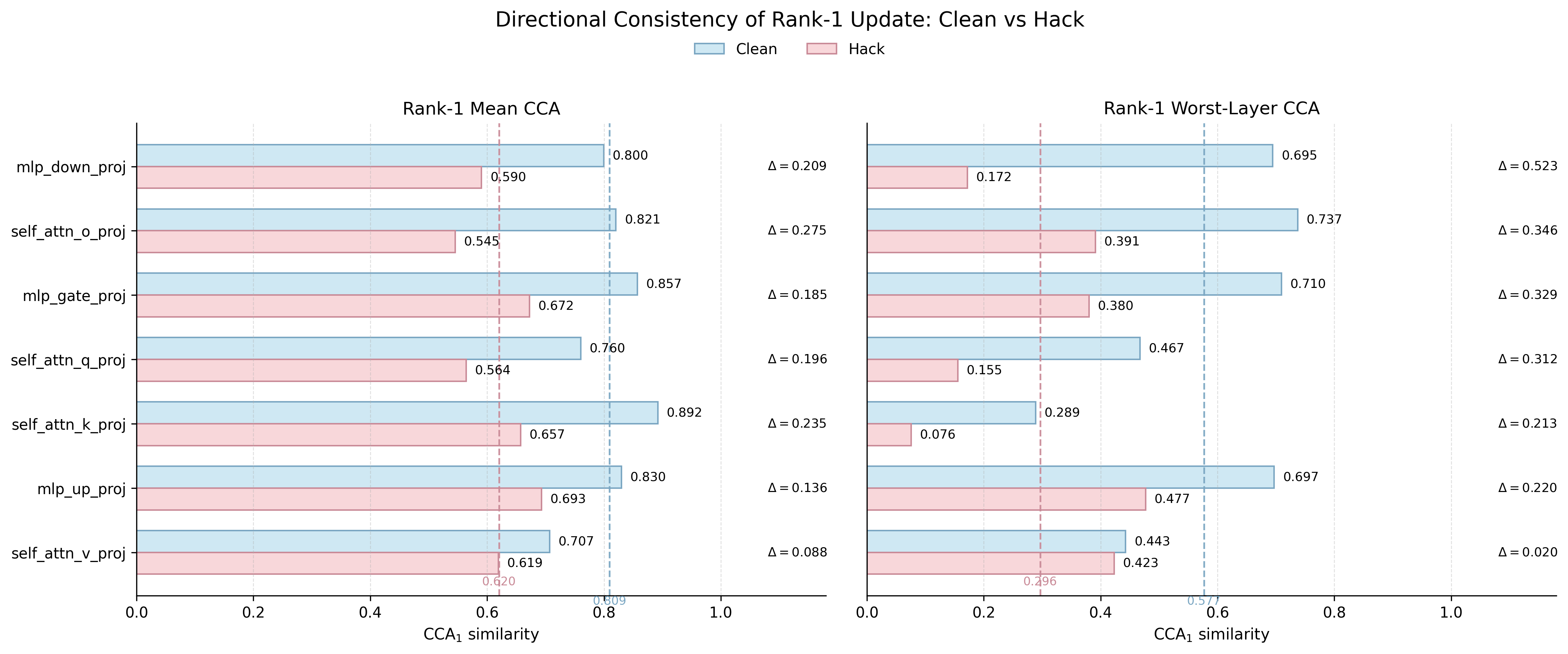}
    \caption{\textbf{Rank-1 CCA similarity of dominant update directions in clean and reward-hacking training.} We compare \(\mathrm{CCA}_1(U_{20}, U_{80})\), the cosine similarity between the dominant update directions at checkpoints \(20\) and \(80\), across different weight matrices. Higher values indicate stronger directional consistency over training. \textbf{Left:} mean rank-1 CCA across layers for each module. \textbf{Right:} worst-layer rank-1 CCA, corresponding to the least aligned layer within each module. Across both views, the reward-hacking model exhibits consistently lower similarity than the clean model, indicating substantially stronger directional drift away from the stable learning trajectory.}
    \label{fig:cca_compare}
    \vspace{-4mm}
\end{figure*}
Let \(W_t\) denote the model parameters at training step \(t\), and define the parameter update as
\begin{align}
    \Delta W_t = W_t - W_0.
\end{align}
We analyze the structure of \(\Delta W_t\) via singular value decomposition (SVD):
\begin{align}
    \Delta W_t = \sum_{i=1}^{r} \sigma_i^{(t)} u_i^{(t)} {v_i^{(t)}}^\top,
\end{align}
where \(\sigma_i^{(t)}\) are the singular values in descending order, and \(u_i^{(t)}\) and \(v_i^{(t)}\) are the corresponding left and right singular vectors.
\begin{definition}[Rank-\(K\) Dominant Direction]
Let \(\Delta W_t = \sum_{i=1}^{r} \sigma_i^{(t)} u_i^{(t)} {v_i^{(t)}}^\top\)
be the singular value decomposition of the parameter update at training step \(t\). We define the \emph{rank-\(K\) dominant update} as the truncated SVD,
\[
    \Delta W_t^{(K)} = \sum_{i=1}^{K} \sigma_i^{(t)} u_i^{(t)} {v_i^{(t)}}^\top.
\]
We further define the corresponding output-direction subspace as
\[
    U_t^{(K)} := \bigl[u_1^{(t)}, \ldots, u_K^{(t)}\bigr],
\]
which represents the principal output-space directions along which the update acts.
\end{definition}
\vspace{-2mm}
\textbf{Interpretation.}
The subspace \(U_t^{(K)}\) captures the dominant modes of change induced by RL training, corresponding to the directions that explain the largest variation in the parameter update. Empirically, these dominant directions account for a substantial portion of the performance gain and evolve smoothly throughout training~\cite{cai2025predictability}. From a functional perspective, the rank-\(K\) update induces the following transformation on a hidden representation \(h\):
\begin{equation}
    \Delta y^{(K)} = \Delta W_t^{(K)} h
    = \sum_{i=1}^{K} \sigma_i^{(t)} u_i^{(t)} \langle v_i^{(t)}, h \rangle.
\end{equation}
This can be interpreted as a superposition of \(K\) key-value operations, where each \(v_i^{(t)}\) selects a relevant input feature and each \(u_i^{(t)}\) determines the corresponding direction of the output update. We focus on controlling the output-direction subspace \(U_t^{(K)}\), while leaving the input-side directions \(\{v_i^{(t)}\}_{i=1}^{K}\) unconstrained.
\\
\textbf{Measuring Directional Change via CCA.}
Given the dominant directions defined above, we quantify how they evolve throughout training. Because dominant update directions may form a low-dimensional subspace when aggregated across layers or checkpoints, we use Canonical Correlation Analysis (CCA) to measure subspace similarity in a geometry-aware manner. For two checkpoints \(t\) and \(s\), let
\begin{equation}
    U_t = \bigl[u_t^{(1)}, \dots, u_t^{(k)}\bigr], \qquad
    U_s = \bigl[u_s^{(1)}, \dots, u_s^{(k)}\bigr], \nonumber
\end{equation}
denote the subspaces spanned by their top-\(k\) singular directions, where \(k\) is the number of retained dominant components. We define their CCA similarity as $
    \mathrm{CCA}_k(U_t, U_s) = \frac{1}{k} \sum_{i=1}^{k} \sigma_i,$
where \(\sigma_i\) are the canonical correlations between the two subspaces. Values close to \(1\) indicate that the two subspaces are highly aligned, whereas smaller values reflect increasingly strong directional drift.
\section{Directional Shift in Reward Hacking}
We analyze how the dominant update direction evolves during training through the rank-1 CCA similarity, \(\mathrm{CCA}_1(U_{20}, U_{80})\), between checkpoints \(20\) and \(80\) (details in \cref{sec:exp}) (Analysis of Rank 5 see~\cref{sec:add_shift}). Larger values indicate stronger directional consistency.
\\
\textbf{Small Direction Shift in Non-Hacking Models.}
We first examine checkpoints that do not exhibit reward hacking. Empirically, their dominant update directions evolve smoothly and consistently over training. As shown in \cref{fig:cca_compare}, the clean run maintains high mean CCA values around $0.8$ across nearly all modules, indicating that the dominant update direction is largely preserved throughout training. Even in the worst-layer view, the similarities remain substantially higher than those of the hacking model, suggesting limited layer-wise drift. Thus, non-hacking training exhibits only a small directional shift.
\\
\textbf{Large Direction Shift in Hacking Models.}
We next examine models that exhibit reward hacking. In contrast to the clean run, the hacking model shows a pronounced loss of directional consistency over training. As shown in \cref{fig:cca_compare}, its mean CCA decreases across nearly all modules by roughly $0.2$, indicating much stronger drift in the dominant update direction. The effect is more severe in the worst-layer view, where several modules reach very low similarity values, sometimes below $0.1$. Overall, the results demonstrate that reward hacking is associated with a substantial departure from the model's intrinsic learning direction.
\section{Method}
Motivated by this observation, we constrain RL updates to the rank-\(K\) dominant subspace, keeping optimization aligned with intrinsic dynamics and preventing drift into hacking directions.
\\
\textbf{Trusted Direction Gradient Alignment.}
\begin{figure*}[th!]
    \centering
    \begin{subfigure}[t]{0.45\linewidth}
        \centering
        \includegraphics[width=\linewidth]{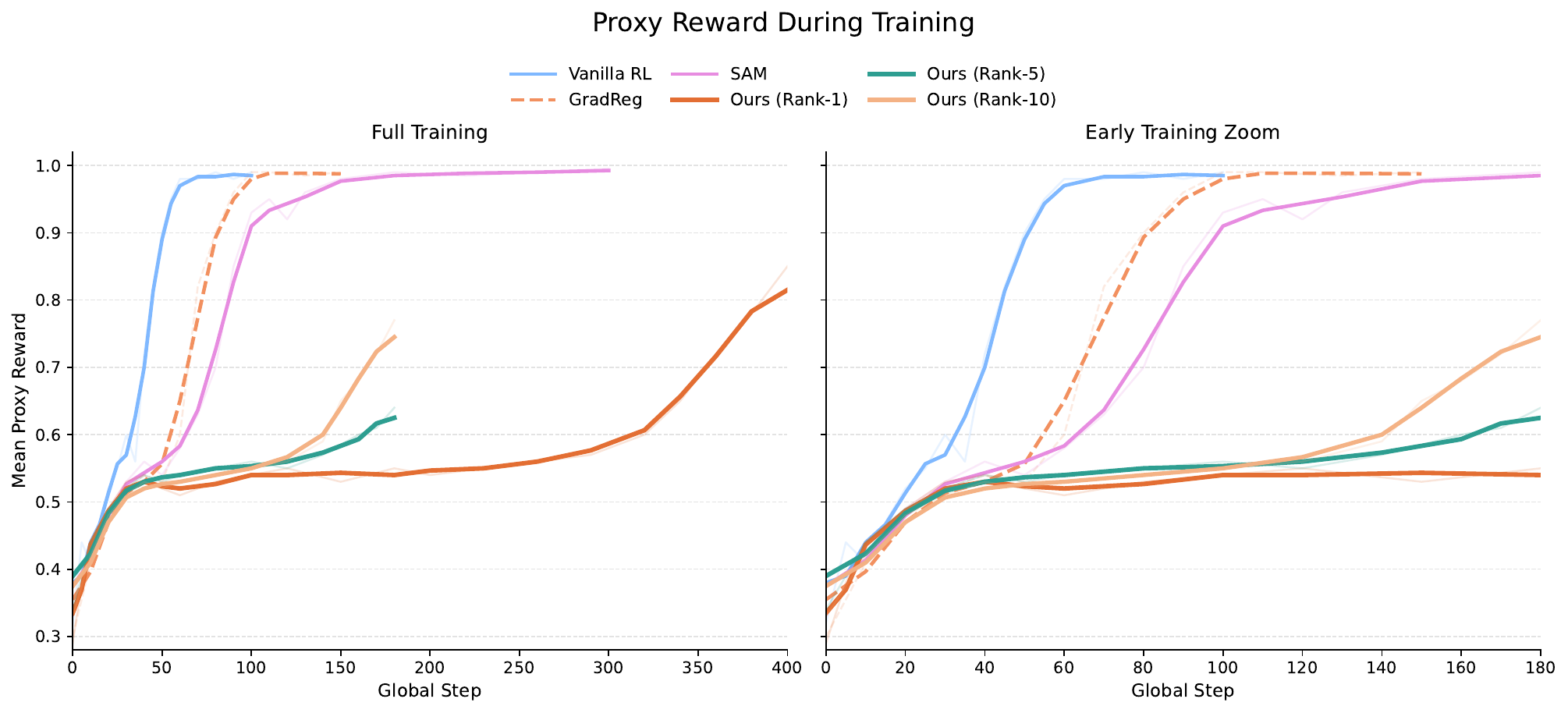}
        \caption{Proxy reward during training.}
        \label{fig:proxy_reward}
    \end{subfigure}
    \hfill
    \begin{subfigure}[t]{0.45\linewidth}
        \centering
        \includegraphics[width=\linewidth]{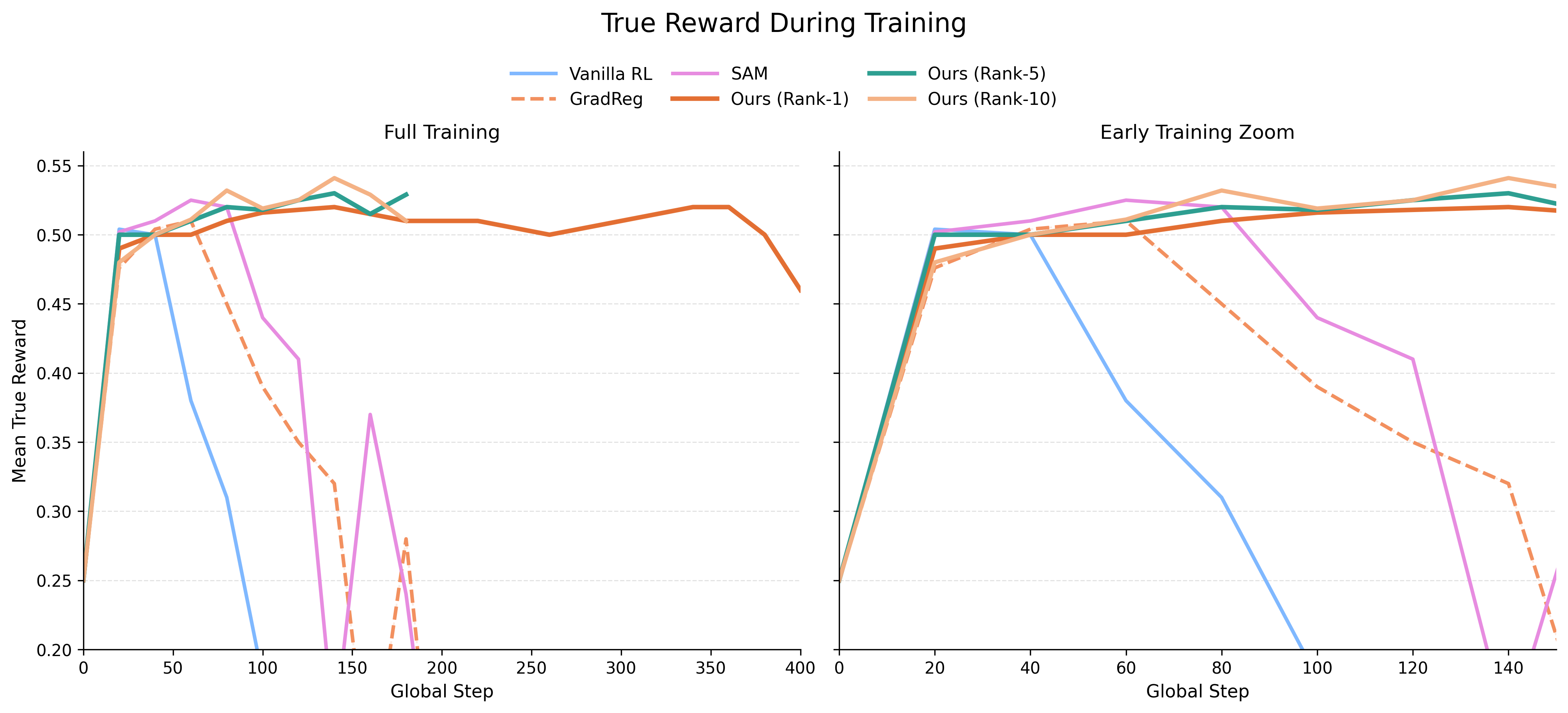}
        \caption{True reward under loophole-free evaluation.}
        \label{fig:true_reward}
    \end{subfigure}
    \caption{\textbf{Delayed reward hacking and improved preservation of true performance.} The left panel shows the evolution of the proxy reward, which reflects the model's ability to exploit the loophole, while the right panel reports the true reward measured under loophole-free evaluation. Faint curves denote raw trajectories and bold curves denote smoothed trends. Compared with vanilla RL, SAM, and gradient regularization, trusted-direction projection slows the rise of proxy reward and maintains substantially more stable true reward throughout training.}
    \vspace{-5mm}
    \label{fig:delay}
\end{figure*}
At training step \(t\), let \(G_t \in \mathbb{R}^{d_{\mathrm{out}} \times d_{\mathrm{in}}}\) denote the gradient of the objective with respect to a model weight matrix. Following \cref{sec:d_direct}, we first estimate a trusted rank-\(K\) output subspace from a short clean-data warmup phase:
\begin{equation}
    U_{\mathrm{clean}}^{(K)} = \bigl[u_1^{\mathrm{clean}}, \dots, u_K^{\mathrm{clean}}\bigr],
\end{equation}
together with the associated singular values \(\{\sigma_i^{\mathrm{clean}}\}_{i=1}^K\). To preserve the relative importance of the dominant clean directions, we define the diagonal weight matrix
\begin{equation}
    \Lambda_{\mathrm{clean}}^{(K)}
    = \operatorname{diag}(\alpha_1, \dots, \alpha_K),
    \qquad
    \alpha_i = \frac{\sigma_i^{\mathrm{clean}}}{\sum_{j=1}^{K} \sigma_j^{\mathrm{clean}}}.
\end{equation}
We then project the gradient onto the trusted output-direction subspace using singular-value weighting:
\begin{equation}
    G_t^{\parallel}
    =
    U_{\mathrm{clean}}^{(K)}
    \Lambda_{\mathrm{clean}}^{(K)}
    {U_{\mathrm{clean}}^{(K)}}^\top
    G_t.
\end{equation}
Our update uses only the trusted component, \(\widetilde{G}_t = G_t^{\parallel}\). This retains the top-\(K\) clean directions and emphasizes each one according to its singular value. As a result, the update remains aligned with the intrinsic clean learning trajectory while suppressing off-subspace components that may introduce instability or encourage reward-hacking behavior.

\section{Experimental Setting}\label{sec:exp}
\textbf{Training Settings.} We follow Wang et al.~\cite{wang2025thinking} and evaluate on Big-Math-RL-Verified under the in-context loophole setting. We use Qwen2.5-3B-Instruct and train on 24,379 examples, with 1,498 examples held out for validation and evaluation. All methods are trained on 8 GPUs with per-device batch size 4, 64 gradient accumulation steps, learning rate \(10^{-5}\), constant scheduling, and KL coefficient \(10^{-3}\). We sample 8 rollouts per prompt during training and 1 during evaluation, with a maximum completion length of 512 tokens. Unless otherwise stated, all methods use the same configuration.
\\
\textbf{Baselines.} We compare against representative RL-stabilization baselines under reward hacking. \textit{Gradient Regularization}~\cite{ackermann2026gradient} smooths optimization by penalizing large or unstable gradients, but mainly controls update magnitude. \textit{SAM}~\cite{foret2020sharpness,kwon2021asam} improves robustness by favoring flatter minima, but targets local loss smoothness. 
\section{Results}
\textbf{Delayed Reward Hacking.}
As shown in \cref{fig:proxy_reward}, vanilla RL rapidly enters the hacking regime, with proxy reward saturating near 0.9 within about 50 steps; gradient regularization and SAM provide only limited delay. In contrast, trusted-direction projection substantially slows saturation: rank-1 does not reach this regime within 400 steps, while rank-5 and rank-10 remain unhacked till 200 steps. Consistently, \cref{fig:true_reward} shows that our methods preserve a higher true reward. \Cref{tab:true_reward_summary} further summarizes this effect by comparing peak, epoch-level true reward across methods.
\begin{table}[t]
    \centering
    \caption{\textbf{Peak and epoch-level true reward.} Values are mean true rewards from the curves in \cref{fig:true_reward}. The peak is computed over the displayed training horizon.}
    \label{tab:true_reward_summary}
    \vspace{1mm}
    \small
    \begin{tabular}{@{}lccc@{}}
        \toprule
        Method & Peak & 1 epoch & 2 epochs \\
        \midrule
        Vanilla RL & \(0.505\) & \(0.190\)   & $0.00$ \\
        GradReg & \(0.509\) & \(0.39\)  & $0.00$\\
        SAM & \(0.525\) & \( 0.44 \)  & $0.00$ \\
        \ours{} (Rank-1) & \(0.522\) & \(0.516\)  & $0.514$\\
        \ours{} (Rank-5) & \(0.530\) & \(0.518\)  & $\mathbf{0.529}$ \\
        \ours{} (Rank-10) & \(\mathbf{0.541}\) & \(\mathbf{0.532}\)  & $0.510$ \\
        \bottomrule
    \end{tabular}
    \vspace{-5mm}
\end{table}
\\
\textbf{Epoch-Level Performance.}
\Cref{tab:true_reward_summary} quantifies the stability gains from trusted-direction projection. While non-\ours{} methods improve early true reward, they collapse by the second epoch, with vanilla RL, gradient regularization, and SAM all falling to 0.000. In contrast, \ours{} improves both peak and long-horizon performance: rank-10 achieves the highest peak at $0.541$ and one-epoch rewards, while rank-5 obtains the best two-epoch value of $0.529$. These results show that trusted-direction projection not only delays proxy-reward saturation but also improves and preserves genuine task performance over longer training.
\\
\textbf{Trade-off with Projection Rank.}
The trusted-subspace rank K controls the trade-off between robustness and flexibility. Smaller ranks enforce stronger alignment with the clean trajectory, suppressing reward hacking more aggressively but limiting adaptation. Larger ranks provide more optimization freedom and better task performance, but weaken the constraint against shortcut directions. As shown in \cref{fig:delay}, rank-1 is the most conservative, while rank-5 and rank-10 better preserve true reward while still delaying hacking relative to baselines. 
\section{Conclusion}
We studied reward hacking in language-model reinforcement learning through the geometry of parameter updates. Our analysis shows that clean training preserves a stable dominant update direction, whereas reward-hacking runs undergo a pronounced directional shift away from this trajectory. Motivated by this finding, we introduced \ours{}, which projects RL gradients onto a trusted subspace estimated from clean supervised updates. Experiments show that \ours{} delays reward hacking and preserves true reward. 
\\
\textbf{Future Work}: We will explore more precise constraints to better unlock model performance while preventing reward hacking. One promising direction, for which we have already observed positive results, is iteratively updating the trusted learning directions. Additional directions are discussed in \cref{sec:future}.

\section*{Acknowledgments} The authors sincerely thank Yida Wang and Xuanqi Zhang for their support. This work was partially funded by the NSERC Discovery Grant RGPIN-2021-03677, Alliance Grant ALLRP 581098-22, the Natural Science and Engineering Research Council of Canada
(NSERC), the Canada CIFAR AI Chairs program, the Canada Research Chair program, an IITP grant funded by MSIT, and the Digital Research Alliance of Canada.

\bibliography{example_paper}

@article{ackermann2026gradient,
  title={Gradient Regularization Prevents Reward Hacking in Reinforcement Learning from Human Feedback and Verifiable Rewards},
  author={Ackermann, Johannes and Noukhovitch, Michael and Ishida, Takashi and Sugiyama, Masashi},
  journal={arXiv preprint arXiv:2602.18037},
  year={2026}
}

@article{cai2025predictability,
  title={On predictability of reinforcement learning dynamics for large language models},
  author={Cai, Yuchen and Cao, Ding and Xu, Xin and Yao, Zijun and Huang, Yuqing and Tan, Zhenyu and Zhang, Benyi and Sun, Guangzhong and Liu, Guiquan and Fang, Junfeng},
  journal={ICLR},
  year={2026}
}

@article{foret2020sharpness,
  title={Sharpness-aware minimization for efficiently improving generalization},
  author={Foret, Pierre and Kleiner, Ariel and Mobahi, Hossein and Neyshabur, Behnam},
  journal={arXiv preprint arXiv:2010.01412},
  year={2020}
}

@article{he2025gardo,
  title={GARDO: Reinforcing Diffusion Models without Reward Hacking},
  author={He, Haoran and Ye, Yuxiao and Liu, Jie and Liang, Jiajun and Wang, Zhiyong and Yuan, Ziyang and Wang, Xintao and Mao, Hangyu and Wan, Pengfei and Pan, Ling},
  journal={arXiv preprint arXiv:2512.24138},
  year={2025}
}

@inproceedings{kwon2021asam,
  title={Asam: Adaptive sharpness-aware minimization for scale-invariant learning of deep neural networks},
  author={Kwon, Jungmin and Kim, Jeongseop and Park, Hyunseo and Choi, In Kwon},
  booktitle={International Conference on Machine Learning},
  pages={5905--5914},
  year={2021},
  organization={PMLR}
}

@article{laidlaw2024correlated,
  title={Correlated proxies: A new definition and improved mitigation for reward hacking},
  author={Laidlaw, Cassidy and Singhal, Shivam and Dragan, Anca},
  journal={arXiv preprint arXiv:2403.03185},
  year={2024}
}

@article{li2025generalist,
  title={Generalist reward models: Found inside large language models},
  author={Li, Yi-Chen and Xu, Tian and Yu, Yang and Zhang, Xuqin and Chen, Xiong-Hui and Ling, Zhongxiang and Chao, Ningjing and Yuan, Lei and Zhou, Zhi-Hua},
  journal={arXiv preprint arXiv:2506.23235},
  year={2025}
}

@article{liu2025inference,
  title={Inference-time scaling for generalist reward modeling},
  author={Liu, Zijun and Wang, Peiyi and Xu, Runxin and Ma, Shirong and Ruan, Chong and Li, Peng and Liu, Yang and Wu, Yu},
  journal={arXiv preprint arXiv:2504.02495},
  year={2025}
}

@article{skalse2022defining,
  title={Defining and characterizing reward gaming},
  author={Skalse, Joar and Howe, Nikolaus and Krasheninnikov, Dmitrii and Krueger, David},
  journal={Advances in Neural Information Processing Systems},
  volume={35},
  pages={9460--9471},
  year={2022}
}

@article{turpin2025teaching,
  title={Teaching models to verbalize reward hacking in chain-of-thought reasoning},
  author={Turpin, Miles and Arditi, Andy and Li, Marvin and Benton, Joe and Michael, Julian},
  journal={arXiv preprint arXiv:2506.22777},
  year={2025}
}

@article{wang2025thinking,
  title={Is It Thinking or Cheating? Detecting Implicit Reward Hacking by Measuring Reasoning Effort},
  author={Wang, Xinpeng and Joshi, Nitish and Plank, Barbara and Angell, Rico and He, He},
  journal={ICLR},
  year={2026}
}

@article{guo2025deepseek,
  title={Deepseek-r1: Incentivizing reasoning capability in llms via reinforcement learning},
  author={Guo, Daya and Yang, Dejian and Zhang, Haowei and Song, Junxiao and Zhang, Ruoyu and Xu, Runxin and Zhu, Qihao and Ma, Shirong and Wang, Peiyi and Bi, Xiao and others},
  journal={arXiv preprint arXiv:2501.12948},
  year={2025}
}

@article{deng2025effect,
  title={On the effect of negative gradient in group relative deep reinforcement optimization},
  author={Deng, Wenlong and Ren, Yi and Li, Muchen and Sutherland, Danica J and Li, Xiaoxiao and Thrampoulidis, Christos},
  journal={arXiv preprint arXiv:2505.18830},
  year={2025}
}

@article{deng2025grpo,
  title={On grpo collapse in search-r1: The lazy likelihood-displacement death spiral},
  author={Deng, Wenlong and Li, Yushu and Gong, Boying and Ren, Yi and Thrampoulidis, Christos and Li, Xiaoxiao},
  journal={arXiv preprint arXiv:2512.04220},
  year={2025}
}

@article{deng2025efficient,
  title={Efficient Forward-Only Data Valuation for Pretrained LLMs and VLMs},
  author={Deng, Wenlong and Zhang, Jiaming and Zeng, Qi and Thrampoulidis, Christos and Gong, Boying and Li, Xiaoxiao},
  journal={arXiv preprint arXiv:2508.10180},
  year={2025}
}
\bibliographystyle{icml2026}
\appendix
\section{Appendix}

\subsection{Future Work}\label{sec:future}
A natural next step is to investigate reward hacking in multi-turn reinforcement learning~\cite{deng2025grpo}. Recent work on inference-time scaling for reward modeling~\cite{liu2025inference} suggests that reward exploitation may become more pronounced in longer-horizon and agentic settings, where a model can exploit the reward through a sequence of intermediate actions rather than a single shortcut. Extending our framework to analyze trajectory-level directional drift may therefore provide a clearer understanding of how reward hacking emerges and accumulates over long-horizon interactions.

Another important direction is to choose the projection rank and training schedule more systematically. Our results suggest a clear trade-off: small ranks suppress hacking more strongly but may over-constrain learning, while larger ranks improve flexibility but weaken robustness. Future work could adapt the rank, RL steps, and clean fine-tuning schedule online using signals such as singular-value decay, directional drift, or validation performance.

\subsection{More Directional Shift} \label{sec:add_shift}
\begin{figure*}[h!]
    \centering
    \includegraphics[width=0.95\linewidth]{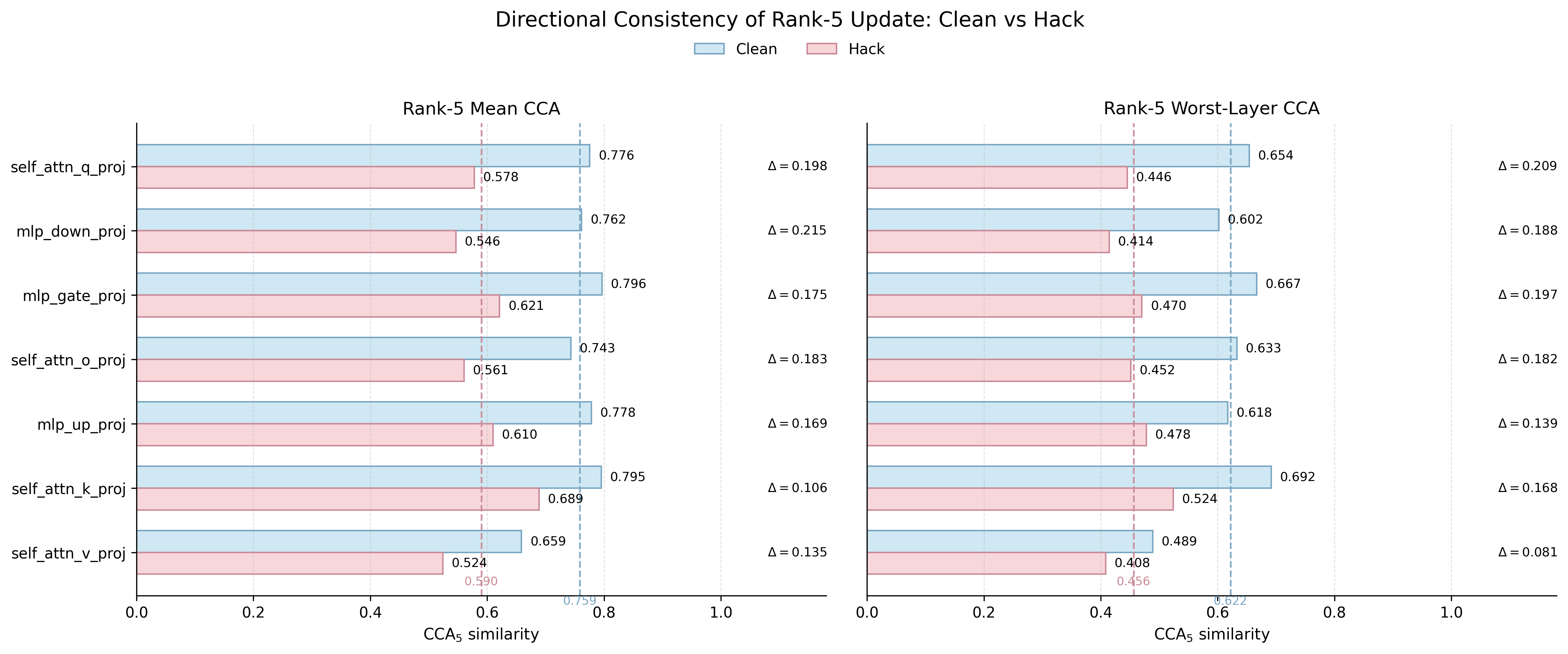}
    \caption{\textbf{Rank-5 CCA similarity of dominant update directions in clean and reward-hacking training.}
    We compare \(\mathrm{CCA}_5(U_{20}, U_{80})\), the similarity between the top-5 dominant update subspaces at checkpoints \(20\) and \(80\), across different weight matrices. Higher values indicate stronger directional consistency over training. \textbf{Left:} mean rank-5 CCA across layers for each module. \textbf{Right:} worst-layer rank-5 CCA, corresponding to the least aligned layer within each module. Across both views, the reward-hacking model exhibits consistently lower similarity than the clean model, indicating stronger directional drift away from the stable learning trajectory.}
    \label{fig:cca_5_compare}
    \vspace{-5mm}
\end{figure*}

Figure~\cref{fig:cca_5_compare} shows that the rank-5 analysis leads to the same qualitative conclusion as the rank-1 result in~\cref{fig:cca_compare}: reward-hacking training deviates more strongly from the clean learning trajectory. Across both the mean-layer and worst-layer views, the hacking run maintains lower CCA similarity than the clean run, indicating that the larger trusted subspace still captures a clear difference in directional stability.

At the same time, moving from rank-1 to rank-5 slightly reduces the absolute CCA values for both clean and hacking runs. This suggests that the approximate linearity of the update trajectory weakens somewhat as additional singular directions are included, since those weaker components are less stable than the leading one. Nevertheless, the clean--hacking gap remains pronounced, showing that the directional-drift phenomenon is robust beyond the single dominant direction.

\subsection*{Impact Statement}
This paper studies reward hacking in reinforcement learning for language models and proposes a mitigation strategy aimed at improving training reliability. Better understanding and controlling reward hacking may reduce unsafe shortcut-seeking behavior in downstream systems, but the same insights could also be used to design stronger proxy objectives or more effective attacks if misapplied.

\end{document}